\documentclass[10pt,twocolumn,letterpaper]{article}

\usepackage[pagenumbers]{iccv} 

\usepackage{multirow}
\usepackage{xcolor,colortbl}

\newcolumntype{s}{>{\columncolor[gray]{.85}[.5\tabcolsep]}c}

\definecolor{iccvblue}{rgb}{0.21,0.49,0.74}
\usepackage[pagebackref,breaklinks,colorlinks,allcolors=iccvblue]{hyperref}

\def\paperID{1717} 
\def\confName{ICCV}
\def\confYear{2025}

\title{Aerial Vision-and-Language Navigation with \\ Grid-based View Selection and Map Construction}

\author{Ganlong Zhao$^{1,2}$ \quad Guanbin Li$^{2,3,4}$\thanks{Corresponding authors are Guanbin Li and Yizhou Yu.} \quad Jia Pan$^{1}$ \quad Yizhou Yu$^{1*}$\\
$^1${The University of Hong Kong} \quad $^2$Sun Yat-sen University\\ 
$^{3}$Guangdong Key Laboratory of Big Data Analysis and Processing \quad $^{4}$Peng Cheng Laboratory\\
{\tt\small  zhaogl@connect.hku.hk, liguanbin@mail.sysu.edu.cn, jpan@cs.hku.hk, yizhouy@acm.org}
}

\begin{document}
\maketitle
\begin{abstract}
Aerial Vision-and-Language Navigation~(Aerial VLN) aims to obtain an unmanned aerial vehicle agent to navigate aerial 3D environments following human instruction.
Compared to ground-based VLN, aerial VLN requires the agent to decide the next action in both horizontal and vertical directions based on the first-person view observations.
Previous methods struggle to perform well due to the longer navigation path, more complicated 3D scenes, and the neglect of the interplay between vertical and horizontal actions.
In this paper, we propose a novel grid-based view selection framework that formulates aerial VLN action prediction as a grid-based view selection task, incorporating vertical action prediction in a manner that accounts for the coupling with horizontal actions, thereby enabling effective altitude adjustments.
We further introduce a grid-based bird's eye view map for aerial space to fuse the visual information in the navigation history, provide contextual scene information, and mitigate the impact of obstacles.
Finally, a cross-modal transformer is adopted to explicitly align the long navigation history with the instruction.
We demonstrate the superiority of our method in extensive experiments.
\end{abstract}    
\section{Introduction}
\label{sec:intro}

Aerial Vision-and-Language Navigation (Aerial VLN)~\cite{liu2023aerialvln} studies Vision-and-Language Navigation (VLN)~\cite{anderson2018vision, ku2020room, qi2020reverie, krantz2020beyond} in aerial environments using unmanned aerial vehicles (UAVs) such as multirotors. Traditional VLN focuses on developing ground agents capable of following human instructions to navigate indoor or outdoor environments.
With the advancement of UAVs, aerial activities have attracted growing interest in numerous research fields.
Many UAV-related applications require aerial agents to comprehend human language for task execution, particularly in scenarios where mission objectives and target locations cannot be specified or planned using GPS signals. For example, drones often rely on language-specified targets or paths to perform tasks, a common requirement in disaster relief operations and other applications involving exploration of unknown environments.
Natural language understanding boosts aerial agents’ interaction with humans during missions, facilitating error correction and tasks like visual question answering while enhancing model reliability in real-world applications.
To address these challenges, Aerial VLN has emerged as a more complex benchmark to evaluate intelligent agents' ability to navigate aerial environments following human instructions.

\begin{figure}
    \centering
    \includegraphics[width=\linewidth]{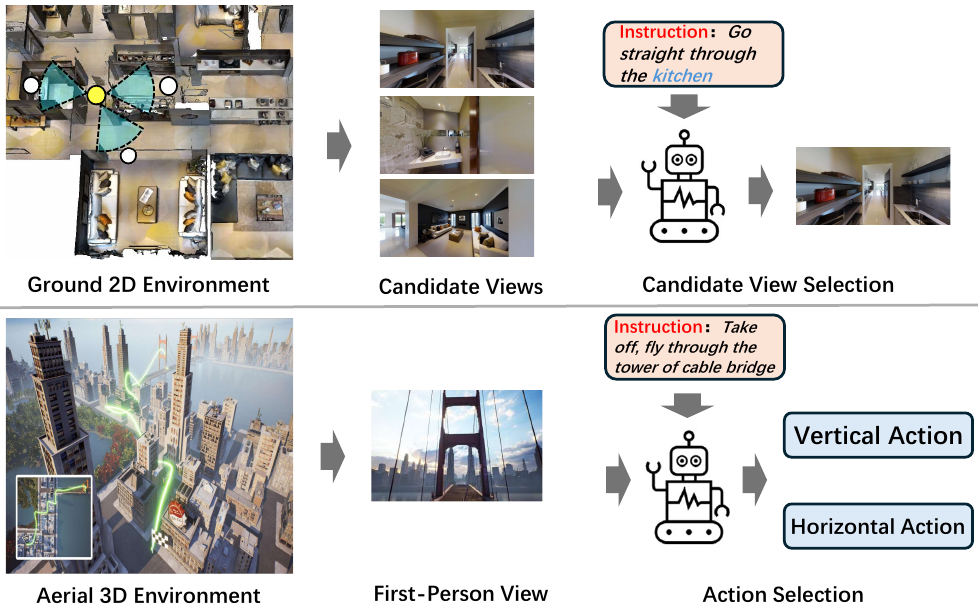}
    \caption{The comparison between aerial VLN and ground-based VLN in discrete environments. Top: The ground-based VLN in discrete environments. The agent receives view candidates at the current position and selects one of them as the next action according to the instruction. Bottom: Aerial VLN. The agent only obtains first-person views and needs to decide both vertical and horizontal actions in continuous environments. }
    \label{fig:teaser}
\end{figure}

Aerial VLN~\cite{liu2023aerialvln} exhibits several key characteristics that make the task particularly challenging~(Fig.~\ref{fig:teaser}). First, aerial VLN introduces the additional vertical dimension in VLN, requiring aerial agents to perform both vertical and horizontal actions. 
Vertical actions are not predicted in isolation by aerial agents; they are intricately coupled with horizontal navigation, as changes in altitude directly impact the field of view and obstacle avoidance.
Ascending increases the camera's viewing range but reduces resolution, while effective obstacle avoidance requires coordinating both horizontal and vertical actions.
Second, aerial VLN often involves navigating complex environments, including many visual obstructions, such as urban areas with intricate three-dimensional structures. The outdoor scenes are further complicated by changing weather and lighting conditions. In expansive and occluded environments, agents may only have a limited view of their surroundings. Successful aerial navigation requires the agent to sufficiently explore and develop a comprehensive understanding of the scene. 
Third, aerial VLN often involves longer instructions and paths than indoor VLN, placing higher demands on the agents' long-term memory and planning capabilities.

Previous research~\cite{liu2023aerialvln} in aerial VLN tries to adapt the Sequence-to-Sequence method~\cite{krantz2020beyond}, which processes a sequence of observations along with an instruction and generates a sequence of actions as output. It failed to achieve satisfactory performance due to the difficulties discussed above.
It ignores the interdependency between vertical and horizontal actions. The agent had to decide on vertical movements solely based on the limited forward-facing visual input, and merging these distinct types of actions, \ie vertical and horizontal, into one prediction stream only further confused the decision-making process. 
Besides, it lacks an explicit mechanism to model historical context and the environment, relying solely on hidden states for state tracking and inference. This simplistic approach becomes insufficient in handling the complex surroundings and extended trajectories in aerial VLN.

As discussed above, it is clear that a thoroughly new framework is required to address the challenges of aerial VLN. 
First, the framework must treat vertical and horizontal action predictions as distinct yet integrated components rather than merging them into a single prediction stream. It should enable the agent to collect and assess information from all directions at its current location for a comprehensive decision.
Second, the scene map specifically designed for aerial environments is necessary, as it enables the agent to leverage longer trajectories to gather comprehensive environmental context, addressing several challenges: limited field of view, maintaining target awareness, backtracking, and navigating complex, obstacle-rich surroundings.
Finally, the complexity of the environment and longer paths require an explicit history construction, along with decision-making that fully leverages a comprehensive understanding of the surroundings.

To this end, we propose a novel approach based on grid-based view selection and map construction. 
First we leverage the elaborate view-based selection and vertical action prediction to enable coordinated decision-making between vertical and horizontal actions in the context of the surrounding environment. Specifically, at each step, the agent surveys its environment by capturing six views—front, back, left, right, up, and down—thus forming a complete skybox. It then selects one view as the horizontal target while predicting the required vertical adjustment (ascend, descend, or maintain altitude) for each view. Additionally, we introduce View-Candidate Correspondence that converts any ground truth path into corresponding ground truth view for training given any agent's position.

For the scene map, we first grid the environment due to its vast size. Then we maintain a top-down grid feature map whose cells record the features of each position in the horizontal plane for aerial environments. We continuously update this feature map in real-time by integrating the agent’s observations along the vertical direction. The agent queries the feature map to extract a local feature map as the scene context to inform decision-making.
Besides, the grid map aligns well with the skybox-based view selection, which further simplifies the feature map updating.

Finally, we employ a cross-modal transformer to match the constructed history embedding sequence with the instruction embedding sequence. It enables more explicit and long-term memory for state tracking, providing the agent with richer and more comprehensive information to support decision-making. In summary, our contributions are:
\begin{itemize}
    \item A grid-based view selection and vertical action prediction framework for aerial VLN that is capable of integrating various VLN methods.
    \item An effective aerial VLN method that leverages BEV feature maps and cross-modal matching to capture rich scene context and ensure precise history alignment.
    \item State-of-the-art performance on two challenging benchmarks for aerial VLN tasks.
\end{itemize}
\section{Related Works}

\noindent \textbf{Aerial Navigation} Unmanned Aerial Vehicle~(UAV) Navigation~\cite{giusti2015machine,smolyanskiy2017toward,majdik2017zurich,shah2018airsim,chen2020valid,liu2023aerialvln} has attracted growing attention in recent years. Earlier methods tackle the problem with GPS or inertial navigation, which is imprecise or can accumulate errors~\cite{laconte2021survey,petritoli2019inertial}. 
Some other methods~\cite{courbon2010vision,lu2018survey,sinopoli2001vision} explore vision-based navigation as an appealing alternative for achieving reliable autonomy.

Aerial Vision-and-Language Navigation incorporates the language modality to aerial navigation.
Some works on quadcopter agents~\cite{DBLP:conf/rss/BlukisBBKA18, blukis2018mapping,misra2018mapping} that follow natural language instructions conduct experiments in a single, small-scale virtual environment with a restricted set of agent actions.
AVDN dataset~\cite{fan2023aerial} focuses on dialogue-based aerial VLN with bird-view image input collected from the satellite images dataset xView~\cite{lam2018xview} and proposes HAA-Transformer for waypoint prediction. It is hard to transfer to our setting where bird-view images are unavailable to the agents. 
\citet{liu2023aerialvln} propose a new Aerial VLN setting and a new benchmark to tackle the problem of VLN with UAVs in aerial environments with first-person view. Aerial VLN~\cite{liu2023aerialvln} collects over 25,000 instructions and 8,446 trajectories in 25 different city-level scenes, and uses Unreal Engine~\cite{UE4} as a 3D simulator to provide continuous environments. 
CityNav~\cite{lee2024citynav} utilizes 3D point cloud data from SensatUrban~\cite{hu2022sensaturban} as UAV flight environments, representing 3D
scans of real-world urban areas, and leverages linguistic annotations and a 3D map with
geographical information from the CityRefer dataset~\cite{miyanishi2023cityrefer} as language-goal information.

\vspace{1mm}
\noindent \textbf{Vision-and-Language Navigation} Vision-and-Language Navigation (VLN)~\cite{anderson2018vision, fried2018speaker, ku2020room, qi2020reverie, wang2019reinforced,cartillier2021semantic,Wangliuyi_2024_CVPR,OVERNAV_2024_CVPR,Wangzihan_2024_CVPR,Liu_2024_CVPR} requires an agent to navigate an environment by following natural language instructions that describe the intended navigation path. 
There are  two major environment settings in VLN benchmarks, discrete~\cite{anderson2018vision,ku2020room,qi2020reverie} and continuous environments~\cite{krantz2020beyond,savva2019habitat}. 
The discrete environments benchmarks require the agent to navigate the environment through a pre-defined viewpoint graph, where the agent teleports to the next viewpoint after selecting it. For continuous environment benchmarks, the agent can move freely within a space without being constrained to fixed locations or paths. 
Due to the superior performance of methods in discrete VLN, some continuous VLN methods~\cite{hong2022bridging,krantz2022sim,an2024etpnav,an20221st} try to tackle the problem by transferring the discrete VLN methods to continuous environments. The key for transfer is the waypoint predictor~\cite{hong2022bridging} that receives panoramic observations and predicts the direction and distance of the next position candidates. However, waypoint predictors rely on pre-defined viewpoints for training and are highly environment-specific and hard to transfer. 
In addition to indoor settings, some VLN datasets focus on more complex outdoor environments, such as the Touchdown dataset~\cite{chen2019touchdown} and a modified version of the LANI dataset~\cite{misra2018mapping}. Some other works~\cite{blukis2019learning} also use drones for instruction-guided navigation.

\section{Method}

\begin{figure*}
    \centering
    \includegraphics[width=0.92\linewidth]{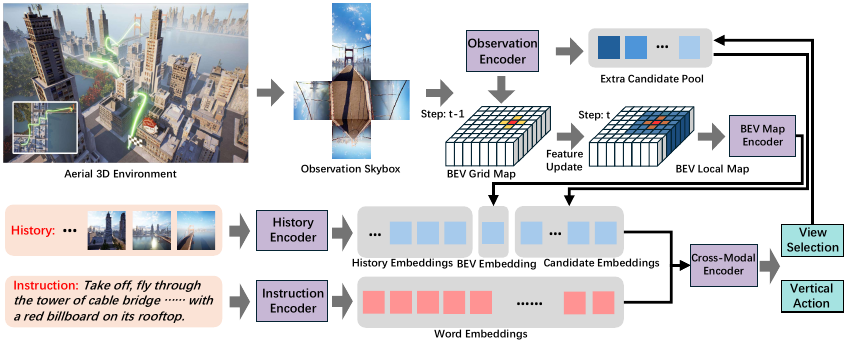}
    \caption{The overall framework of our method. The agent receives observations skybox at the current position in aerial 3D environments, and the BEV grid map uses the observations to update the cell features. The BEV local map is sent to the BEV map encoder to generate the BEV embedding to provide the context of surrounding environments. The observation encoder encodes the observations along with the extra candidates as candidate embeddings. The agent extracts the instruction embeddings and history embeddings with the corresponding encoder, and the concatenated visual embeddings and word embeddings are sent to the cross-modal encoder for view selection and vertical action. The final prediction is then used to update the extra candidate pool. The figure of aerial 3D environment is from \cite{liu2023aerialvln}.}
    \label{fig:framework}
\end{figure*}

The framework of our proposed method is depicted in Fig.~\ref{fig:framework}. In the upper part, the agent receives the observation skybox at the current position in aerial 3D environments and prepares the candidate features and BEV grid map features for model input. In the lower part, the model encodes the history and instruction with corresponding encoders and uses a cross-modal encoder for view selection and vertical action prediction. The prediction results are then used to update the extra candidate pool.

\subsection{Problem Formulation}

Given an instruction $I = \{w_1, w_2, ... w_{N_I}\}$ of $N_I$ words, an aerial agent starts at the initial position $p_0 = (x_0, y_0, z_0)$ with an initialized orientation 
in a 3D environment. The aerial agent needs to navigate this environment following the instruction by performing actions from the available action set $A = \{$\textit{Move Forward, Turn Left, Turn Right, Ascend, Descend, Move Left, Move Right, Stop}$\}$. In AerialVLN and AerialVLN-S~\cite{liu2023aerialvln}, the vertical actions move the agent 2 units, while horizontal actions move it 5 units. The turning action is executed in increments of 15 degrees. At each step $t$, the agent receives the environment observation $O_t$ and decides the next action $a_t$. When the agent stops or reaches the maximum step number, the navigation is regarded as a success when the agent is within 20 meters of the ground truth destination. 

\subsection{Aerial VLN by Grid-based View Selection}
Previous aerial VLN research restricts the agent's decision-making to what it can see directly ahead. This restriction can undermine the reliability of the agent's decisions, for example, losing track of the target when the instructions lack explicit directional cues or when the agent's orientation does not match the intended path. Moreover, previous studies~\cite{hong2022bridging} have shown that view selection is an effective strategy in ground-based VLN, 
but their solutions cannot be applied to aerial VLN, because view selection in ground-based VLN requires a waypoint predictor~\cite{hong2022bridging} which is unavailable in aerial environments.
In this section, we propose a grid-based view selection framework that successfully transforms the action prediction in aerial VLN to view selection in the skybox observations without the waypoint predictor. The proposed framework bridges the gap between previous VLN research and aerial VLN, allowing for the integration of various VLN methods and serving as a foundation for future research.

\vspace{1mm}
\noindent \textbf{View-Candidate Correspondence} Given the agent’s current position at step $t$, $p_t = (x_t, y_t, z_t)$, we obtain an observation skybox $O_t = \{o_t^i\}_{i=1}^6$, including observation images from the agent’s front, left, right, back, top, and bottom views, respectively. This can be accomplished by equipping the agent with three cameras (front, up, and down) and having the agent rotate 360 degrees horizontally.
For each image $o_t^i$ in $O_t$, we assign a candidate position $\tilde{p}_t^i$ based on the image direction and the agent's step length:
\begin{equation}
    \tilde{p}_t^i = (x_t, y_t, z_t) + \mathbf{d_t^i} \odot (s_h, s_h, s_v),
    \label{eq:view-candidate}
\end{equation}
where $(x_t, y_t, z_t)$ is the position of agent and cameras, $s_h$ is the agent's horizontal step length and $s_v$ is the vertical step length. $\mathbf{d_t^i}$ is the normalized orientation vector of the camera that captures observation $o_t^i$. $\odot$ is the element-wise product. Thus the resulting $\tilde{p}_t^i$ indicates the position where the agent will be after moving one step toward the direction that the camera is oriented.

It is important to note that the assignment of candidate positions can be generalized to any environmental observations based on the camera position $p_t$ and orientation $\mathbf{d_t^i}$. Therefore, the selection of views can be transformed into a choice among position candidates.

\vspace{1mm}
\noindent \textbf{Ground Truth Candidate Generation} When the agent is located at position $p_t$ and receives $N$ views and their corresponding position candidates $\{\tilde{p}_t^i\}_{i=1}^N$, the training process requires a ground truth candidate for loss calculation. To generate this ground truth candidate, we compared the ground truth path $P_{gt}$ and the navigation path $P_t$ up to step $t$ using normalized Dynamic Time Warping~(nDTW):
\begin{equation}
    c_{gt} = \arg \max_{1 \leq i \leq N} \{ \text{nDTW}(P_{gt}, (P_t + \tilde{p}_t^i))\},
    \label{eq:gt_candidate}
\end{equation}
where $(P_t + \tilde{p}_t^i)$ represents the navigation path after moving to the candidate $\tilde{p}_t^i$, and $\textbf{nDTW}$ compute the path similarity between $P_{gt}$ and $(P_t + \tilde{p}_t^i)$. The index $c_{gt}$ is the generated ground truth candidate at step $t$ for further processing.

To ensure the local precision of the ground truth candidate and to avoid deviations from the local path due to it being too close to the endpoint, we set $P_{gt}$ as a sub-path of the entire ground truth path. Specifically, $P_{gt}$ starts at $p_0$ and ends at $p_{t + L}^{gt}$, where $p_t^{gt}$ is the point on the ground truth path closest to the agent's current position $p_t$, and $p_{t + L}^{gt}$ represents the position reached by moving $L$ steps from $p_t^{gt}$ towards the endpoint along the ground truth path, which is the $(t + L)$-th element.

\vspace{1mm}
\noindent \textbf{Low-level Control} Given the current position $p_t$ and the selected candidate position $\tilde{p}_t^i$, we need to move the agent to the target position. The movement is performed horizontally and vertically respectively. More specifically, we repeat the following procedure:
\begin{itemize}
    \item Turn the agent until the relative heading between the agent's orientation and the vector from $p_t$ to $\tilde{p}_t^i$ for the horizontal direction is less than half the turning angle.
    \item Move the agent forward for one step if the horizontal distance between $p_t$ and $\tilde{p}_t^i$ is larger than half the horizontal step length.
    \item Move the agent up/down for one step if the vertical distance between $p_t$ and $\tilde{p}_t^i$ is larger than half the vertical step length.
    \item update $p_t$ to the new position.
\end{itemize}
The procedure is repeated until the relative distance between $p_t$ and $\tilde{p}_t^i$ is less than half the step length for horizontal and vertical directions simultaneously. The \textit{Move Left} and \textit{Move Right} actions are deprecated in this procedure. 
If there’s an obstacle in the path of the action, the agent will maneuver around it by moving upward to avoid it.

\vspace{1mm}
\noindent \textbf{Grid-based View Selection} 
The above three processes do not rely on a scene grid, but we still employ a grid-based approach to simplify the framework and to align with the BEV grid map construction in the next section. Specifically, we capture six views from the agent’s front, back, left, right, top, and bottom perspectives at each step. The angles between the four horizontal views are multiples of 90 degrees. Thus, selecting any of these views will result in the agent's rotation being in multiples of 90 degrees. This approach effectively discretizes the scene based on the agent’s initial position and orientation, with grid cells defined by the agent’s step size in each direction. Grid-based scene discretization simplifies low-level control and scene map construction, allowing the agent to navigate between grid vertices and plan its movements accordingly.

\subsection{Model Design}

The grid-based view selection framework allows traditional VLN methods to be adapted for aerial environments. However, we need to develop several important designs and modules to address the specific challenges of aerial VLN. In this section, we introduce a novel method for aerial VLN based on History-Aware Multimodal Transformer (HAMT)\cite{chen2021history}, a ground-based VLN approach for discrete environments. We provide a detailed overview of our model architecture, training methodology, and essential design modifications.~(Fig.~\ref{fig:framework}).

\vspace{1mm}
\noindent \textbf{Model Architecture} Given the instruction $I = \{w_1, w_2, ... w_{N_I}\}$ of $N_I$ words, the text encoding process first uses BERT~\cite{DBLP:conf/naacl/DevlinCLT19} to extract the instruction embeddings, then employs a transformer~\cite{vaswani2017attention} to obtain the contextual representation $X = (x_{CLS}, x_1, x_2, ... x_{N_I})$. For the view list $\mathcal{O}_t = \{o_1, ... o_{M_t}\}$ at step $t$ with $M_t$ available views, observation encoding process first computes the relative angles, including relative heading $\theta_i$ and relative elevation $\phi_i$ for each view $o_i$, then extracts the observation embedding $e_i$ as the sum of the view embedding and the angle embeddings. The resulting observation embeddings $E_t = (e_1, ... e_{M_t}, e_{stop})$ includes all $M_t$ view embeddings and a \textit{stop} embedding. 

The hierarchical history encoding further encodes the navigation history $\mathcal{H}_t$ at step $t$, which consists of all past panoramic observations $\mathcal{O}_{1, ..., t-1}$ and executed actions $a_{1, ..., t-1}$. A panoramic transformer encodes the panoramic observations to produce panoramic embeddings, then a temporal transformer is used to capture the temporal relations between them and produces the history embeddings $H_t = (h_{CLS}, h_1, ..., h_{t-1})$. 
The history and observation embeddings are concatenated to form the visual modality, while the instruction embedding represents the textual modality. These embeddings are sent to the cross-modal transformer, which outputs a probability distribution over different views of the observation $O_t$, determining the predicted action.

\vspace{1mm}
\noindent \textbf{Vertical Action Prediction}
In contrast to traditional VLN, a key challenge in aerial VLN is to predict the agent's actions in the vertical direction. We address this challenge through two primary design choices. First, as mentioned earlier, the agent receives six different views at each step, including upward and downward perspectives, and selects one to determine its next position. The choice of upward or downward views allows the agent to adjust its vertical position, either ascending or descending.
Second, we let the agent predict a vertical offset $d_v^i \in [0, 1]$ for each view $o_i$ in the \textbf{horizontal} direction, providing two extra position candidates for both upward and downward movements besides the candidate $\tilde{p}_t^i$ in Eq.~\ref{eq:view-candidate}. 
Specifically, for a given view $o_i$ and its corresponding position candidate $\tilde{p}_t^i$, we first calculate a set of three candidate positions: lower, middle, and upper:
\begin{equation}
    U_i = (\tilde{p}_t^i - v_{up}, \tilde{p}_t^i, \tilde{p}_t^i + v_{up}),
\end{equation}
where $v_{up}$ is the displacement vector corresponding to an upward movement. Then similar to Eq~\ref{eq:gt_candidate}, we compute the ground truth vertical offset for each observation $o_i$ by selecting the candidate with the highest nDTW score:
\begin{equation}
    d_{v_{gt}}^i = \arg\max_{1 \leq j \leq 3} \{\text{nDTW}(P_{gt}, (P_t + \tilde{p}_j))\}, \tilde{p}_j \in U_i.
\end{equation}
Finally, we use Mean Square Error~(MSE) to calculate the loss for each view:
\begin{equation}
    L_{vertical} = \frac{1}{N}\sum_{i=1}^N \text{MSE}(d_v^i, (d_{v_{gt}}^i - 1) / 2),
\end{equation}
where $N$ is the number of observation candidates.

For inference, the agent predicts the vertical offset $d_v$ for each view and chooses an observation $o_i$ as the next action. We then select the corresponding candidate position in $U_i$ according to $d_v^i$ as the target position for low-level control.

\vspace{1mm}
\noindent \textbf{BEV Grid Map}
Due to the complexity of aerial scenes, particularly in urban environments where numerous buildings obstruct views, we propose a Bird's Eye View~(BEV) grid map based on feature fusion. This map is designed to provide the agent with local spatial information for action decision-making. Given the agent's initial position, we construct a two-dimensional grid map $M$ centered on that position, with the grid size determined by the agent's horizontal step length. Each cell in the grid map contains the BEV feature vector for the corresponding location.

For any given view $o_t^i$ and its position candidate $\tilde{p}_t^i$ at step $t$, we extract the feature of the view using the agent's feature extractor. We then identify the nearest cell $(u, v)$ in the grid map to $\tilde{p}_t^i$. Finally, we update the features of this cell using a GRU:
\begin{equation}
    M_{(u, v)}^t = \text{GRU}(M_{(u, v)}^{t-1}, f_{\tilde{p}_t^i}),
\end{equation}
where $M_{(u, v)}^{t}$ is the feature in $M$ at $(u, v)$ after step $t$, and $f_{\tilde{p}_t^i}$ is the extracted feature. Next, we extract a local grid feature map $M_{local}$ of size $L_M \times L_M$ and send it to the agent. The agent uses a BEV local map encoder to extract the embedding $e_{bev}$.
The BEV local map embedding $e_{bev}$ serves as the vision-modal context, encapsulating visual information about the surrounding environment. We integrate $e_{bev}$ between the history embeddings and the observation embeddings as vision-modal input for the cross-modal transformer to predict the next action.

\begin{table*}[!tbp]
\centering
{
\resizebox{0.95\linewidth}{!}{%
\begin{tabular}{p{0.3cm}<{\centering}p{2.5cm}|ccccc|ccccc}
\hline
& \multicolumn{1}{c|}{} & \multicolumn{5}{c|}{\textbf{Validation Seen}}   & \multicolumn{5}{c}{\textbf{Validation Unseen}} \\
\multirow{-2}{*}{\textbf{\#}} & \multicolumn{1}{c|}{\multirow{-2}{*}{\textbf{AerialVLN}}} & NE/m $\downarrow$ & SR/\% $\uparrow$ & OSR/\% $\uparrow$ & SDTW/\% $\uparrow$ & NDTW/\% $\uparrow$ & NE/m $\downarrow$ & SR/\% $\uparrow$ & OSR/\% $\uparrow$ & SDTW/\% $\uparrow$ & NDTW/\% $\uparrow$ \\ 
\hline
\rowcolor[HTML]{EFEFEF} 
1                             & Random                                                    & 300.8             & 0.0              & 0.0               & 0.0       &    -     & 351.0             & 0.0              & 0.0               & 0.0     &     -      \\
2                             & Action Sampling                                           & 383.1             & 0.1              & 2.1               & 0.1       &    -     & 434.9             & 0.2              & 2.1               & 0.1     &     -      \\
\rowcolor[HTML]{EFEFEF} 
3                             & Seq2Seq                                                   & 480.4             & 2.9              & 10.2              & 1.0      &    -      & 551.8             & 1.1              & 5.6               & 0.3    &     -       \\
4                             & CMA                                                       & 293.5             & 2.3              & 6.5               & 0.8      &    -      & 360.7             & 1.6              & 4.4               & 0.5    &     -       \\
\rowcolor[HTML]{EFEFEF} 
5                             & Ours                                                     & \textbf{271.1}                 & \textbf{7.5}                & \textbf{12.6}                 & \textbf{3.3}    & \textbf{11.7}         & \textbf{333.5}                 & \textbf{3.2}                & \textbf{8.1}                 & \textbf{1.3}        & \textbf{7.9}          \\ \hline
\end{tabular}
}
}
\caption{Performance of baselines and our method on the AerialVLN Dataset. We evaluate the performance of our method across four metrics commonly used in previous studies for comparison. Additionally, we report the nDTW score. Our method demonstrates significant improvements across all metrics, with a more pronounced improvement on the validation seen set compared to the validation unseen set.
}
\label{tab:aerialvln}
\end{table*}

\begin{table*}[!tbp]
\centering
{
\resizebox{0.95\linewidth}{!}{%
\begin{tabular}{p{0.3cm}<{\centering}p{2.5cm}|ccccc|ccccc}
\hline
& \multicolumn{1}{c|}{} & \multicolumn{5}{c|}{\textbf{Validation Seen}}   & \multicolumn{5}{c}{\textbf{Validation Unseen}} \\
\multirow{-2}{*}{\textbf{\#}} & \multicolumn{1}{c|}{\multirow{-2}{*}{\textbf{AerialVLN-S}}} & NE/m $\downarrow$ & SR/\% $\uparrow$ & OSR/\% $\uparrow$ & SDTW/\% $\uparrow$ & NDTW/\% $\uparrow$ & NE/m $\downarrow$ & SR/\% $\uparrow$ & OSR/\% $\uparrow$ & SDTW/\% $\uparrow$ & NDTW/\% $\uparrow$ \\ 
\hline
\rowcolor[HTML]{EFEFEF} 
S1                            & Random                                                      & 109.6             & 0.0              & 0.0               & 0.0    &-            & 149.7             & 0.0              & 0.0               & 0.0  &-              \\
S2                            & Action Sampling                                             & 213.8             & 0.9              & 5.7               & 0.3    &-            & 237.6             & 0.2              & 1.1               & 0.1  &-              \\
\rowcolor[HTML]{EFEFEF} 
S3                            & LingUNet                                                 & 383.8             & 0.6              & 6.9              & 0.2        &-        & 368.4             & 0.4              & 3.6              & 0.9       &-         \\
S4                            & Seq2Seq                                                     & 146.0             & 4.8              & 19.8              & 1.6    &-            & 218.9             & 2.3              & 11.7              & 0.7  &-              \\
\rowcolor[HTML]{EFEFEF} 
S5                            & CMA                                                         & 121.0             & 3.0              & 23.2              & 0.6    &-            & 172.1             & 3.2              & 16.0              & 1.1  &-               \\
S6                            & Seq2Seq-DA                                                  & 85.5              & 9.9              & 24.1              & 4.5    &-            & 143.5             & 4.0              & 10.9              & 0.7  &-               \\
\rowcolor[HTML]{EFEFEF} 
S7                            & CMA-DA                                                      & 92.2              & 9.9              & 26.5              & 3.7    &-            & 122.7             & 4.5              & 13.9              & 1.0  &-               \\
S8                            & LAG                                               & 90.2              & 7.2              & 15.7              & 2.4              &-  & 127.9             & 5.1              & 10.5              & 1.4            &-    \\ 
\rowcolor[HTML]{EFEFEF} 
S9                             & Ours                                                     & \textbf{70.3}                 & \textbf{20.8}                & \textbf{33.4}                 & \textbf{10.2}        & \textbf{27.9}          & \textbf{121.3}                 & \textbf{7.4}                & \textbf{16.1}                 & \textbf{2.5}         & \textbf{13.7}         \\ \hline
\end{tabular}
}
}
\caption{Performance of baselines and our method on the AerialVLN-S Dataset. We evaluate our method on five metrics, demonstrating significant improvements over all competitors on four of them. The improvement is more significant on AerialVLN-S than AerialVLN. 
}
\label{tab:aerialvln-s}
\end{table*}

\vspace{1mm}
\noindent \textbf{Extra Candidates}
It is important to note that in grid-based view selection, the agent makes action decisions by choosing from the available views. At each position, the agent can obtain six different views. The grid-based view selection does not impose any constraints on the number or location of views. We introduce an extra candidate pool to store candidate views from the navigation history that have high confidence but were not selected. When the agent reaches a new position, the six new views are processed alongside the views in the extra candidate pool for prediction. These prediction results are then used to update the extra candidate pool. 
Specifically, we implement a straightforward approach by setting the size of the extra candidate pool to $s_{pool}$, and we add the top $s_{pool}$ samples with the highest confidence from each model prediction to the pool, filtering out candidates that have already been visited. The inclusion of extra candidates helps the model correct navigation errors and prevents it from getting lost due to the complexity and occlusions in the environment.

\section{Experiment}

Following previous studies~\cite{liu2023aerialvln}, we present the experiment results on AerialVLN and AerialVLN-S in this section.

\vspace{1mm}
\noindent \textbf{Dataset} AerialVLN~\cite{liu2023aerialvln} is a challenging benchmark for aerial VLN in continuous environments. 
It requires agents to navigate with a first-person view in city-level scenarios.
AerialVLN contains four different dataset splits: train, validation seen, validation unseen and test. The training/test set contains 16,380/4,830 instructions from 17/8 scenes, while the val-seen/val-unseen contains 1,818/2,310 instructions from 17/8 scenes. 
AerialVLN-S~\cite{liu2023aerialvln} is a variant of AerialVLN for small scenes. It retains the same data split but includes 17 scenes that are smaller in scale and feature evenly distributed path lengths, leading to a reduced overall path length. The training/test/val-seen/val-unseen contains 10,113/771/333/531 instructions from 12/5/12/5 scenes.
We do not provide the experiment result on the test set as the evaluation server for test set has not been released.

\vspace{1mm}
\noindent \textbf{Evaluation Metric} Following previous studies~\cite{liu2023aerialvln}, we report the performance of our method using four metrics: Navigation Error~(NE), the distance between the stopping location to the destination; Success Rate~(SR), the ratio of agents that stop within 20 meters of the destination; Oracle Success Rate~(OSR), the ratio of navigation episodes in which the agent has been within 20 meters of the destination at any point; Success rate weighted by Normalised Dynamic Time Warping~(SDTW), which weights the similarity between ground truth path and navigation path by success rate. We further provide the Normalised Dynamic Time Warping~(nDTW) score for a detailed and comprehensive comparison, which quantifies the overall similarity between the navigation path and the ground truth path.

\vspace{1mm}
\noindent \textbf{Implementation Details} We implement our method based on \cite{chen2021history}, and retain most of the hyper-parameters and model architectures. 
We use CLIP ViT-B/16~\cite{radford2021learning} model for image feature extraction. 
The batch size is set to 8 in all experiments.
The learning rate is set to 0.0001. The weight for vertical loss $L_{vertical}$ is 1. 
The size of BEV local map $L_M$ is set to 11, and the number of extra candidates is 10.
$L$ is set as 5.
All experiments are conducted on four RTX-3090 GPUs with 24GB memory, and we run the AirSim simulator~\cite{shah2018airsim} on two other GPUs. The models for both AerialVLN and AerialVLN-S are trained for 60000 iterations.

\vspace{1mm}
\noindent \textbf{Experiment Result} The experimental results on AerialVLN and AerialVLN-S are shown in Table~\ref{tab:aerialvln} and Table~\ref{tab:aerialvln-s}. In Table~\ref{tab:aerialvln}, we compare our method to four different baselines: Random, Action Sampling, Seq2Seq~\cite{krantz2020beyond}, and CMA~\cite{krantz2020beyond}. In Random, the agent randomly selects actions during navigation until the \textit{Stop} action is selected or the max step number is reached. Action Sampling model samples the actions according to the action distribution of the training set. Seq2Seq and CMA are two continuous VLN baselines that predict actions given the observation and instruction inputs. In Table~\ref{tab:aerialvln-s}, we compare our method to LingUNet~\cite{misra2018mapping}, a baseline proposed for LANI~\cite{misra2018mapping}, Seq2Seq-DA, CMA-DA and LAG~\cite{liu2023aerialvln}. ``-DA" indicates the method adopts Dataset Aggregation~\cite{ross2011reduction} technique in training, and LAG~\cite{liu2023aerialvln} is the previous best method. The performance of other methods in both tables are copied from \cite{liu2023aerialvln}.

As shown in Table~\ref{tab:aerialvln}, our method achieves significantly better performance in all metrics. Compared to Random and Action Sampling, Seq2Seq achieves a higher success rate (both SR and OSR), but its navigation error~(NE) is significantly larger. This suggests that the higher success rate of Seq2Seq is partly due to its longer navigation paths, which increase the likelihood of the agent being closer to the destination during the navigation. CMA outperforms Random and Action Sampling in NE and SR, but its performance is still far from satisfying. Our method achieves 4.6\% improvement in success rate while also reducing navigation error.
On validation unseen set, our method also shows significant improvement across all metrics, demonstrating its effectiveness in unfamiliar environments.

In Table~\ref{tab:aerialvln-s}, we evaluate our method on AerialVLN-S, which contains the instructions in smaller scenes. Despite the addition of more competitors, our method still achieves the best performance. The relatively simpler tasks in AerialVLN-S allow our method's advantages to become even more evident. Our method achieves a 10.9\% improvement in success rate on the val-seen split, more than doubling the performance of the previous best method while also reducing navigation error. On the val-unseen split, our method improves the success rate by 2.3\%, again with smaller navigation error. In both tables, the performance improvements on the val-seen split are greater than those on the val-unseen split, indicating that prior knowledge of the scene plays a crucial role in complex aerial VLN tasks, particularly in long-distance navigation.

\begin{figure*}
    \centering
    \includegraphics[width=\linewidth]{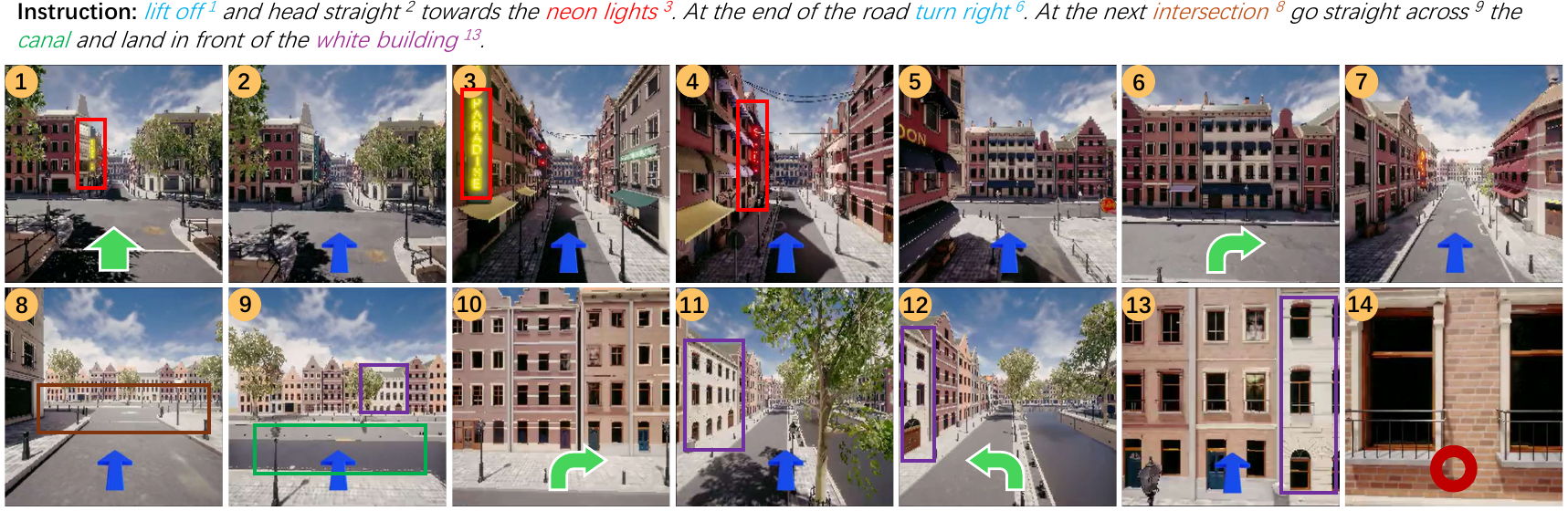}
    \caption{Visualization of a successful navigation of our method. Blue arrows indicate the \textit{Forward} action; green arrows represent the vertical and turning actions~(\textit{Turn Left, Turn Right, Ascending, Descending}). The final red circle denotes Stop. We highlight aligned landmarks by colored bounding boxes in images and words in the instruction using the same color. The superscript of words denotes the index of the corresponding action or object in images.}
    \label{fig:visualization}
\end{figure*}

We also present a qualitative result in Fig.~\ref{fig:visualization}. It shows that our method can successfully identify the landmarks and follow the directional instructions for navigation.

\section{Ablation Study}

\subsection{Ablation on Different Components}

To demonstrate the effectiveness of each component, we ablate them to provide a more comprehensive study in Table~\ref{tab:ablation_studies}.

We ablate four components, \textit{Extra Candidate~(EC), BEV Grid Map~(BGM), Top/Down Observation~(TDO), and Vertical Action~(VA)} and show the experimental results on AerialVLN-S validation seen in Table~\ref{tab:ablation_studies}. For \textit{Extra Candidate}, we remove the component by setting the number of extra candidates to zero. \textit{Top/Down Observation}, we remove the upward and downward observations during training and evaluation to eliminate the context information. In this case, the agent can only move vertically by the vertical action prediction.  This does not affect the agent's ability to reach the ground truth destination, because the distance threshold for success~(20m) is much higher than the step size~(5m). For \textit{Vertical Action}, we remove the vertical action prediction in both training and evaluation, thus the agent can only move up or down by selecting the corresponding view. Top/Down Observation and Vertical Action can not be ablated simultaneously because the agent needs to move vertically in the navigation.
As shown in Table~\ref{tab:ablation_studies}, all four components contribute to performance improvement.

\begin{table}[!tbp]
    \resizebox{\linewidth}{!}{%
		\begin{tabular}{c|cccc|cccc}
			\hline
			\scriptsize \shortstack{\#} &
			{\footnotesize \textsc{EC}}
                & \footnotesize \textsc{BGM}
			& \footnotesize \textsc{TDO}
                & \footnotesize \textsc{VA}
			& {NE/m $\downarrow$}
			& {SR/\% $\uparrow$}
			& {OSR/\% $\uparrow$}
			& {SDTW/\% $\uparrow$}
			\\
			\midrule
                \scriptsize \texttt{1}
                & &
                & \checkmark &
                & 74.8
                & 14.5
                & 28.0
                 & 6.3
			\\
			\scriptsize \texttt{2}
                & &
                & & \checkmark
                & 74.9
                & 15.7
                & 28.3
                 & 7.0
			\\
                \scriptsize \texttt{3}
			    & &
                & \checkmark & \checkmark
                & 75.7
                & 16.3
                & 29.2
                & 6.9
			\\
                \scriptsize \texttt{4}
			    & & \checkmark
                & \checkmark & \checkmark
                & 72.2
                & 18.7
                & 29.8
                 & 8.2
			\\
                \scriptsize \texttt{5}
			    & \checkmark & \checkmark
                & \checkmark & \checkmark
                & 70.3
                & 20.8
                & 33.4
                & 10.2
			\\
			\bottomrule
		\end{tabular}
        }
	\caption{The ablation of different components in our method on validation seen of AerialVLN-S. \textit{EC, BGM, TDO} and \textit{VA} represent \textit{Extra Candidate, BEV Grid Map, Top/Down Observation} and \textit{Vertical Action} respectively. All four components contribute to the performance improvement.}
	\label{tab:ablation_studies}
\end{table}

\begin{table}[!tbp]
    \resizebox{\linewidth}{!}{%
		\begin{tabular}{c|l|cccc}
			\hline
			\scriptsize \shortstack{\#} &
			{\footnotesize \textsc{Model}}
			& {NE/m $\downarrow$}
			& {SR/\% $\uparrow$}
			& {OSR/\% $\uparrow$}
			& {SDTW/\% $\uparrow$}
			\\
			\midrule
			\scriptsize \texttt{1}
                & Ours~($L_M = 7$)
                & 70.3
                & 19.1
                & 28.6
                 & 9.5
			\\
                \scriptsize \texttt{2}
			    & Ours~($L_M = 9$)
                & 70.0
                & 19.9
                & 29.8
                & 9.1
			\\
                \scriptsize \texttt{3}
			    & Ours~($L_M = 11$)
                & 70.3
                & 20.8
                & 33.4
                 & 10.2
			\\
                \scriptsize \texttt{4}
			    & Ours~($L_M = 13$)
                & 67.7
                & 20.8
                & 31.1
                & 10.1
			\\
			\bottomrule
			\midrule
			\scriptsize \texttt{1}
                & Ours~($s_{pool} = 5$)
                & 70.5
                & 20.2
                & 31.2
                & 8.3
			\\
                \scriptsize \texttt{2}
			    & Ours~($s_{pool} = 10$)
                & 70.3
                & 20.8
                & 33.4
                & 10.2
			\\
                \scriptsize \texttt{3}
			    & Ours~($s_{pool} = 15$)
                & 69.4
                & 19.6
                & 35.8
                & 9.7
			\\
                \scriptsize \texttt{4}
			    & Ours~($s_{pool}=20$)
                & 69.6
                & 19.3
                & 34.1
                & 9.2
			\\
                \bottomrule
		\end{tabular}
        }
	\caption{The performance of our method on AerialVLN-S validation seen with different $L_M$ and $s_{pool}$. $L_M$ decides the size of the local BEV grid map sent to the agent and $s_{pool}$ is the number of candidates maintained in the candidate pool for view selections. Our method achieves the best success rate with $L_M = 11$ and $s_{pool} = 10$. Our method outperforms competitors in all experiments.}
	\label{tab:sensitivity}
\end{table}

\subsection{Sensitivity for Hyper-Parameter $L_M$ and $s_{pool}$}

Here we analyze the sensitivity of two important hyper-parameters, the size of the local BEV grid map $L_M$ and the number of extra candidates $s_{pool}$. $L_M$ decides the range of local map information sent to the agent to provide contexts for action prediction. $L_M$ is set to an odd number because the local map is agent-centralized. $s_{pool}$ decides the number of extra candidates maintained in the candidate pool to provide additional choices for action prediction.

The experiment results on AerialVLN-S validation seen are shown in Table~\ref{tab:sensitivity}. The best success rate of our method is achieved with $L_M = 11$ and $s_{pool} = 10$ and in all experiments our method outperforms other competitors. Both hyper-parameters do not greatly affect the performance. The extra candidate pool does not always contain the correct view for the next action and can only help the action prediction when the agent makes mistakes. 
Besides, the oracle success rate increases with higher $s_{pool}$, given that the agent is more likely to navigate a larger scene area.

\subsection{Vertical Action Prediction Analysis}

\begin{table}[!tbp]
    \resizebox{\linewidth}{!}{%
		\begin{tabular}{c|l|ccccc}
			\hline
                &
			& \multicolumn{2}{c}{{Select View}}
		      &
                & \multicolumn{2}{c}{{All Views}}
                \\
                \cmidrule{3-4}
			\cmidrule{6-7}
			\scriptsize \shortstack{\#} &
			{\footnotesize \textsc{Model}}
			& {Exact Acc}
			& {Relaxed Acc}
                &
			& {Exact Acc}
			& {Relaxed Acc}
			\\
			\midrule
			\scriptsize \texttt{1}
                & Train
                & 52.9
                & 98.9
                &
                & 69.6
                 & 98.8
			\\
                \scriptsize \texttt{2}
			    & Val Seen
                & 54.3
                & 98.9
                &
                & 70.0
                & 99.1
			\\
                \scriptsize \texttt{3}
			    & Val Unseen
                & 45.1
                & 99.0
                &
                & 68.3
                 & 98.7
			\\
			\bottomrule
		\end{tabular}
        }
	\caption{The performance of vertical action prediction in our method on AerialVLN-S. ``Select View'' is the accuracy on the views selected by the agent as the next action, while ``All Views'' is the accuracy calculated on all views sent to the agent for view selection. ``Exact Acc'' is the accuracy of vertical action prediction while ``Relaxed Acc'' counts correct if the prediction and ground truth differ by at most 1. Our method significantly outperforms random selection in all settings.}
	\label{tab:vertical_prediction_analysis}
\end{table}

Here we evaluate the performance of an important component in our method, vertical action prediction, and show the experiment results in Table~\ref{tab:vertical_prediction_analysis}. Given $N$ view candidates at step $t$, the agent outputs one view as the action prediction and $N$ vertical actions for all views. The vertical action can take one of three values: -1 for downward, +1 for upward and 0 for maintaining height. As shown in Table~\ref{tab:vertical_prediction_analysis}, we test our method on the three splits of AerialVLN-S and reports the performance on four different metrics.  ``Select View'' indicates the accuracy on the view selected by the agent as the next action, while ``All Views'' is the accuracy calculated on all views sent to the agent for view selection. Besides, the ``Exact Acc'' is the accuracy that the prediction matches ground truth exactly, while the ``Relaxed Acc'' counts correct if prediction and ground truth differ by at most 1. 

The vertical prediction is a three-class classification problem, thus the accuracy of random selection is about 33\%. Our method significantly outperforms random selection in all settings. The Performance on ``All Views'' is much higher than ``Select View'', which may be because the selected view is more likely to be at the boundary of moving upward/downward. The ``Relaxed Acc'' is close to 100\% in all settings, indicating that the possibility that the agent moves further from the ground truth height is very low.

\section{Conclusion}

We propose a grid-based view selection and map construction method for aerial Vision-and-Language Navigation. Grid-based view selection aims to formulate the aerial VLN in continuous environments as the view selection task in discrete environments. The map construction further fuses the observation features along the navigation path to construct a top-down grid map to provide information on surrounding environments. 
Extensive experiments on AerialVLN and AerialVLN-S demonstrate that the grid-based view selection is an effective framework that adapts traditional VLN methods to aerial VLN, and the BEV grid map enables the agent to utilize the environment context for better
performance. 
{
    \small
    \bibliographystyle{ieeenat_fullname}
    \bibliography{main}
}

\end{document}